%% file: main.tex
\documentclass[11pt,a4paper]{article}
\usepackage{times,latexsym}
\usepackage{url}
\usepackage[T1]{fontenc}

\usepackage[acceptedWithA]{tacl2018v2}

\usepackage{graphicx}
\usepackage{tabularx}
\usepackage{subcaption}

\usepackage{amsmath}

\usepackage[utf8]{inputenc} 

\title{Discriminating Between Similar Nordic Languages}

\author{René Haas\\
 IT University of Copenhagen \\
  {\sf renha@itu.dk} \\ \And
  Leon Derczynski\\
 IT University of Copenhagen \\
  {\sf leod@itu.dk} 
}

\begin{document}
\maketitle

\begin{abstract}
Automatic language identification is a challenging problem. Discriminating between closely related languages is especially difficult. This paper presents a machine learning approach for automatic language identification for the Nordic languages, which often suffer miscategorisation by existing state-of-the-art tools. Concretely we will focus on discrimination between six Nordic languages: Danish, Swedish, Norwegian (Nynorsk), Norwegian (Bokmål), Faroese and Icelandic.
\end{abstract}

\input{sec/01introduction}
\input{sec/02related-work}

\input{sec/03data}
\input{sec/04baselines}

\input{sec/05our-approach}

\input{sec/06results}

\input{sec/07analysis}
\input{sec/08conclusion}

\section{References}
\label{main:ref}

\bibliographystyle{acl_natbib}
\bibliography{references}

\end{document}

%% file: sec/01introduction.tex
\section{Introduction}
Automatic language identification is a challenging problem and especially discriminating between closely related languages is one of the main bottlenecks of state-of-the-art language identification systems~\cite{DSL2014}.

Language technology for Scandinavian languages is in a nascent phase (e.g.~\newcite{kirkedal2019lacunae}). One problem is acquiring enough text with which to train e.g. large language models. Good quality language ID is critical to this data sourcing, though leading models often confuse similar Nordic languages.

This paper presents a machine learning approach for automatic language identification between six closely-related Nordic languages: Danish, Swedish, Norwegian (Nynorsk), Norwegian (Bokmål), Faroese and Icelandic.

This paper explores different ways of extracting features from a corpus of raw text data consisting of Wikipedia summaries in respective languages and evaluates the performance of a selection of machine learning models.

Concretely we will compare the performance of classic machine learning models such as Logistic Regression, Naive Bayes, Support vector machine, and K nearest Neighbors with more contemporary neural network approaches such as Multilayer Perceptrons (MLP) and Convolutional Neural Networks (CNNs).

After evaluating these models on the Wikipedia data set we will continue to evaluate the best models on a data set from a different domain in order to investigate how well the models generalize when classifying sentences from a different domain.

%% file: sec/02related-work.tex
\section{Related Work}

The problem of discriminating between similar languages has been investigated in recent work \cite{DSLEvaluation,DSL2015} which discuss the results from two editions of the ``Discriminating between Similar Languages (DSL) shared task". Over the two editions of the DSL shared task different teams competed to develop the best machine learning algorithms to discriminate between the languages in a corpus consisting of 20K sentences in each of the languages: Bosnian, Croatian, Serbian, Indonesian, Malaysian, Czech, Slovak, Brazil Portuguese, European Portuguese, Argentine Spanish, Peninsular Spanish, Bulgarian and Macedonian.

%% file: sec/03data.tex
\section{The Nordic DSL Dataset}
This section describes the construction of the Nordic DSL (Distinguishing Similar Lanugages) dataset.

Data was scraped from Wikipedia. We  downloaded summaries for randomly chosen Wikipedia articles in each of the languages, saved as raw text to six {\tt .txt} files of about 10MB each. While Bornholmsk would be a welcome addition~\cite{derczynski2019bornholmsk}, exhibiting some similarity to Faroese and Danish, there is not yet enough digital text.

After the initial cleaning (described in the next section) the dataset contained just over 50K sentences in each of the language categories. From this, two datasets with exactly 10K and 50K sentences respectively were drawn from the raw dataset. In this way the datasets are stratified, containing the same number of sentences for each language.

We split these datasets, reserving 80\% for the training set and 20\% for the test set.

\subsection{Data Cleaning}
This section describes how the dataset is initially cleaned and how sentences are extracted from the raw data.

\paragraph{Extracting Sentences}

The first pass in sentence tokenisation is splitting by line breaks.
We then extract shorter sentences with the sentence tokenizer ({\tt sent\_tokenize}) function from the NLTK\cite{nltk} python package. This does a better job than just splitting by {\tt '.'} due to the fact that abbreviations, which can appear in a legitimate sentence, typically include a period symbol.

\paragraph{Cleaning characters}
The initial dataset has many characters that do not belong to the alphabets of the languages we work with. Often the Wikipedia pages for people or places contain names in foreign languages. For example a summary might contain Chinese or Russian characters which are not strong signals for the purpose of discriminating between the target languages.

Further, it can be that some characters in the target languages are mis-encoded. These mis-encodings are also not likely to be intrinsically strong or stable signals.

To simplify feature extraction, and to reduce the size of the vocabulary, the raw data is converted to lowercase and stripped of all characters with are not part of the standard alphabet of the six languages.

In this way we only accept the following character set
\begin{verbatim}
'abcdefghijklmnopqr
stuvwxyzáäåæéíðóöøúýþ '
\end{verbatim}

and replace everything else with white space before continuing to extract the features.
For example the raw sentence
\begin{verbatim}
'Hesbjerg er dannet ved
sammenlægning af de 2 gårde
Store Hesbjerg
og Lille Hesbjerg i 1822.'
\end{verbatim}
will be reduced to
\begin{verbatim}
'hesbjerg er dannet ved sammenlægning
 af de gårde store hesbjerg
 og lille hesbjerg i ',
\end{verbatim}
We thus make the assumption that capitalisation, numbers and characters outside this character set do not contribute much information relevant for language classification.

%% file: sec/04baselines.tex
\section{Baselines}

\subsection{Baseline With langid.py}

As a baseline to compare the performance of the models in we compare with an off-the-shelf language identification system. ``langid.py: An Off-the-shelf Language Identification Tool." \cite{langID} is such a tool.

{\tt langid.py} comes with with a pretrained model which covers 97 languages. The data for langid.py comes from from five different domains: government documents, software documentation, newswire, online encyclopedia and an internet crawl. Features are selected for cross-domain stability using the LD heuristic~\cite{lui2011cross}.

We evaluated how well langid.py performed on the Nordic DSL data set. It is a peculiar feature of the Norwegian language that there exist two different written languages but three different language codes. Since langid.py also returned the language id ``no" (Norwegian) on some of the data points we restrict langid.py to only be able to return either ``nn" (Nynorsk) or ``nb" (Bokmål) as predictions.

\begin{figure}
  \centering
  \includegraphics[width = 200pt]{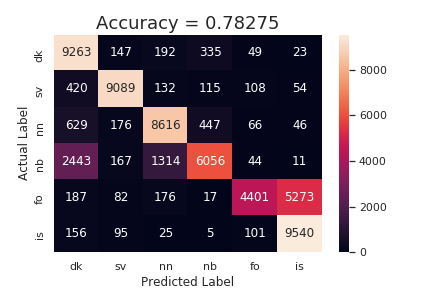}
  \caption{Confusion matrix with results from langid.py on the full dataset, 300K instances.}
  \label{langid_confusion_matrix}
\end{figure}

Figure~\ref{langid_confusion_matrix} shows the confusion matrix for the langid.py classifier. The largest confusions were between Danish and Bokmål, and between Faroese and Icelandic. We see that langid.py was able to correctly classify most of the Danish instances; however, approximately a quarter of the instance in Bokmål were incorrectly classified as Danish and just under an eighth was misclassified as Nynorsk.

Furthermore, langid.py correctly classified most of the Icelandic data points; however, over half of the data points in Faroese were incorrectly classified as Icelandic.

\begin{table}
  \centering
  \begin{tabular}{ l | c | r }
    \hline
    Model               & Encoding  & Accuracy \\
    \hline
    Knn                 & cbow &  0.780\\
    Log-Reg             & cbow &  0.819\\
    Naive Bayes         & cbow &  0.660\\
    SVM                 & cbow &  0.843\\
    Knn                 & skipgram &  0.918\\
    Log-Reg             & skipgram &  \textbf{0.929}\\
    Naive Bayes         & skipgram &  0.840\\
    SVM                 & skipgram &  \textbf{0.928}\\
    Knn                 & char bi-gram  & 0.745\\
    Log-Reg             & char bi-gram  & 0.907\\
    Naive Bayes         & char bi-gram  & 0.653\\
    SVM                 & char bi-gram  & 0.905\\
    Knn                 & char uni-gram  & 0.620\\
    Log-Reg             & char uni-gram  & 0.755\\
    Naive Bayes         & char uni-gram  & 0.614\\
    SVM                 & char uni-gram  & 0.707\\
    \hline
  \end{tabular}
  \caption{Overview of results for the data set with 10K data points in each language.}
  \label{baseline-results-10k}
\end{table}

\subsection{Baseline with linear models}

Table~\ref{baseline-results-10k} shows results for running the models on a data set with 10K sentences in each language category. We see that the models tend to perform better if we use character bi-grams instead of single characters.

Also we see that logistic regression and support vector machines outperform Naive Bayes and K-nearest neighbors in all cases. Furthermore, for all models, we get the best performance if we use the skip-gram model from FastText.

Comparing the CBOW mode from FastText with character bi-grams we see that the CBOW model is on par with bi-grams for the KNN and Naive Bayes classifiers, while bi-grams outperform  CBOW for Logistic Regression and support vector machines.

%% file: sec/05our-approach.tex
\section{Our Approach}

\subsection{Using FastText}

The methods described above are quite simple. We also compared the above method with FastText, which is a library for creating word embeddings developed by Facebook~\cite{BagOfTricks}.

\newcite{EnrichingWordVectors} explain how FastText extracts feature vectors from raw text data. FastText makes word embeddings using one of two model architectures: continuous bag of words (CBOW) or the continuous skip-gram model.

The skip-gram and CBOW models are first proposed in~\cite{EfficientWordRepresentations} which is the paper introducing the word2vec model for word embeddings. FastText builds upon this work by proposing an extension to the skip-gram model which takes into account sub-word information.

Both models use a neural network to learn word embedding from using a context windows consisting of the words surrounding the current target word. The CBOW architecture predicts the current word based on the context, and the skip-gram predicts surrounding words given the current word~\cite{EfficientWordRepresentations}.

\begin{figure}
  \centering
  \includegraphics[width = 200pt]{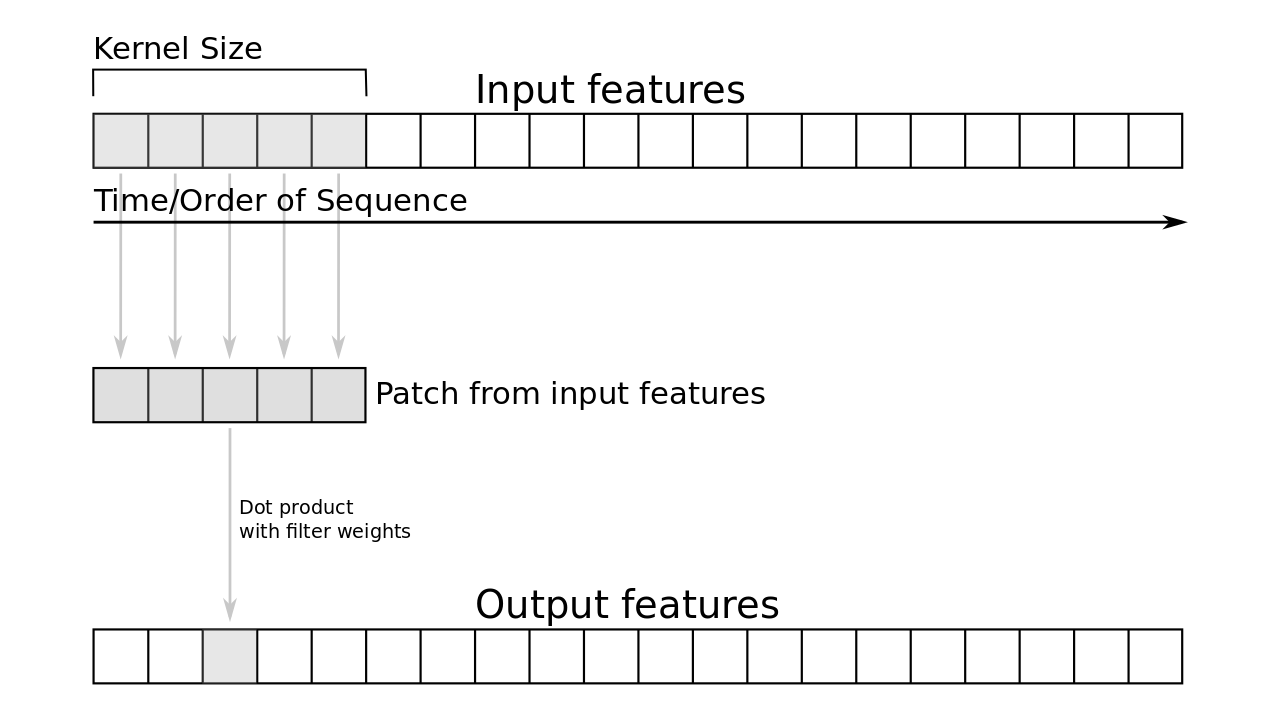}
  \caption{Diagram of Convolutional Neural network.}
  \label{cnn}
\end{figure}

\subsection{Using A Convolutional Neural Network}

While every layer in a classic multilayer perceptron is densely connected, such that each of the nodes in a layer are connected to all nodes in the next layer, in a convolutional neural network we use one or more convolutional layers. Convolutional Neural Networks have an established use for text classification~\cite{textcnn_google}.

The basic premise of a convolutional layer is illustrated in Figure~\ref{cnn}.\footnote{Source: \url{https://realpython.com/python-keras-text-classification/}} In a CNN a filter ``slides" over the input. The CNN then takes e.g. the dot product of the weights of the filter and the corresponding input features, before applying a further function.

%% file: sec/06results.tex
\begin{table}
  \centering
  \begin{tabular}{ l | c | r }
    \hline
    Model               & Encoding  & Accuracy \\
    \hline
    MLP                 & char bi-gram &  0.898 \\
    CNN                 & char bi-gram & \textbf{0.956} \\
    MLP                 & char uni-gram &  0.697\\
    CNN                 & char uni-gram  & 0.942 \\
    \hline
  \end{tabular}
  \caption{Overview of results for the neural network models for the data set with 10K data points in each language.}
  \label{keras-results}
\end{table}

\section{Results}

\subsection{Results with neural networks}
Results for the neural network architectures are in Table~\ref{keras-results}. Here we compare the result of doing character level uni- and bi-grams using Multilayer Perceptron and Convolutional neural networks. We see that the CNN performs the best, achieving an accuracy of 95.6\% when using character bi-grams. Both models perform better using bi-grams than individual characters as features while the relative increase in performance is greater for the MLP model.

\subsection{Increasing the size of the data set}
Often the performance of supervised classification models increases with more training data. To measure this effect we increase the amount of training data to 50K sentences in each of the language categories. Due to longer training times only the baseline models were included, with the skip-gram encoding from FastText which we saw achieved the highest accuracy.

\begin{table}
  \centering
  \begin{tabular}{ l  c | r }
    \hline
    Model               & Encoding & Accuracy \\
    \hline
    Knn                 & skipgram & 0.931\\
    Logistic Regression & skipgram  & \textbf{0.933}\\
    Naive Bayes         & skipgram  & 0.806\\
    SVM                 & skipgram& 0.925\\
    \hline
  \end{tabular}
  \caption{Overview of results for the data set with 50K data points in each language.}
  \label{results-sklearn300k}
\end{table}

\begin{table}
  \centering
  \begin{tabular}{ l c | r }
    \hline
    Model               & Encoding & Accuracy \\
    \hline
    MLP                 & char bi-gram  & 0.918\\
    CNN                 & char bi-gram  & \textbf{0.970}\\
    \hline
  \end{tabular}
  \caption{Overview of results for the data set with 50K data points in each language.}
  \label{results-keras-300k}
\end{table}

\begin{figure}
  \centering
  \includegraphics[scale=0.5]{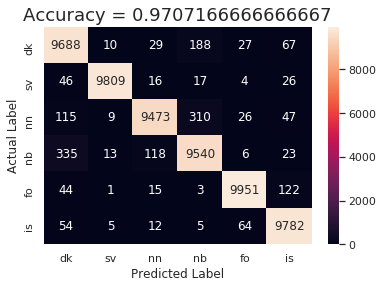}
  \caption{Confusion matrix with results from the convolutional neural network on the full data set with 50K data points in each language.}
  \label{confusion_matrix-big-cnn}
\end{figure}

Table~\ref{results-sklearn300k} shows that the performance of the logistic regression model and the K-nearest-neighbors algorithm improved slightly by including more data. Unexpectedly, performance of the support vector machine and Na\"{i}ve Bayes dropped slightly with extra data.

Even when including five times the amount of data, the best result, logistic regression with an accuracy of 93.3\%, is still worse than for the Convolutional Neural Network trained on 10K data points in each language.

In Table~\ref{results-keras-300k} we see the results for running the neural networks on the larger data set. Both models improve by increasing the amount of data and the Convolutional Neural Network reached an accuracy of 97\% which is the best so far.



In Figure~\ref{confusion_matrix-big-cnn} we see the confusion matrix for the convolutional Neural Network trained on the full Wikipedia data set with 300K data points pr language. We see that the largest classification errors still happens between Danish, Bokmål and Nynorsk as well as between Icelandic and Faroese.

\begin{figure}
  \centering
  \includegraphics[scale=0.5]{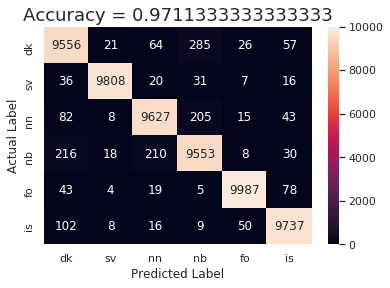}
  \caption{Confusion matrix with results from a supervised FastText model on the full data set with 300K data points.}
  \label{fasttext_supervised}
\end{figure}

\subsection{Using FastText supervised}

FastText can also be used for be supervised classification. In~\newcite{BagOfTricks} the authors show that FastText can obtain performance on par with methods inspired by deep learning, while being much faster on a selection of different tasks, e.g. tag prediction and sentiment analysis. We apply FastText classification to the Nordic DSL task. The confusion matrix from running the FastText supervised classifier can be seen in Figure~\ref{fasttext_supervised}. We see that FastText performance is similar to that of the CNN.

\subsection{Cross-domain evaluation}

Training on single-domain data can lead to classifiers that only work well on a single domain. To see how the two best performing models generalize, we tested on a non-Wikipedia data set.

For this, we used Tatoeba,\footnote{{\tt tatoeba.org/}} a large database of user-provided sentences and translations. 

The language style used in the Tatoeba data set is different from the language used in Wikipedia. The Tatoeba data set mainly consists of sentences written in everyday language. Below we see some examples from the Danish part of the Tatoeba data set.

\begin{quote}
Hvordan har du det? (How are you?)

På trods af al sin rigdom og berømmelse, er han ulykkelig. (Despite all his riches and renown, he is unlucky.)

Vi fløj over Atlanterhavet. (We flew over the Atlantic Ocean.)

Jeg kan ikke lide æg. (I don't like eggs.)

Folk som ikke synes at latin er det smukkeste sprog, har intet forstået. (People who don't think Latin is the most beautiful language have understood nothing.)
\end{quote}



\begin{figure}
    \centering
    \begin{subfigure}[b]{0.45\textwidth}
      \includegraphics[scale=0.5]{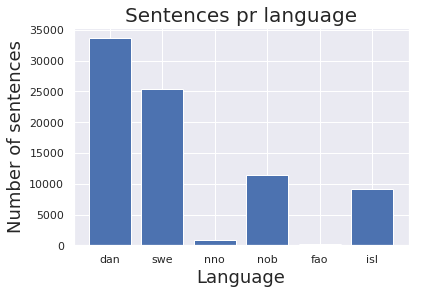}
      \caption{Distribution of the number of sentences in each language in the Tatoeba data set.}
      \label{tatoebasentprlang}
    \end{subfigure}
    ~
    \begin{subfigure}[b]{0.45\textwidth}
      \includegraphics[scale=0.5]{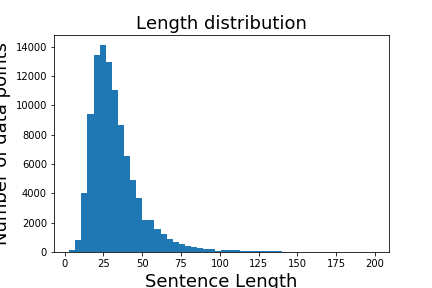}
      \caption{Distribution of the length of sentences in the Tatoeba data set.}
      \label{tatoebalengths}
    \end{subfigure}
    \caption{Distribution of the lengths and language classes of Tatoeba sentences.}
\end{figure}

In Figure~\ref{tatoebasentprlang} we see the number of sentences in each language for all sentences in the Tatoeba data set. Observe that we have very few samples in Nynorsk and Faroese.

We see that the performance drops when shifting to Tatoeba conversations. For reference the accuracy of langid.py on this data set is 80.9\% so FastText actually performs worse than the baseline with an accuracy of 75.5\% while the CNN is better than the baseline with an accuracy of 83.8 \%.

\begin{figure}
    \centering
    \includegraphics[scale=0.5]{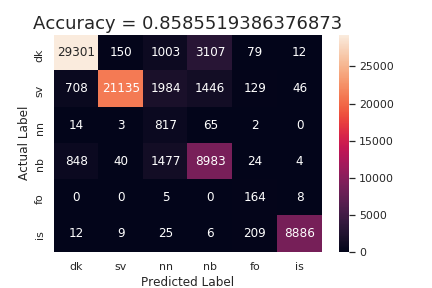}
    \caption{Confusion matrix for FastText trained using only character level n-grams on the Wikipedia data set and evaluated on the Tatoeba data set.}
    \label{fasttextcharngram}
\end{figure}

One explanation for the drop in performance is that the sentences in the Tatoeba data are significantly shorter than the sentences in the Wikipedia data set as seen in Figure~\ref{tatoebalengths}. As we saw in the previous section, both models tend to mis-classify shorter sentences more often than longer sentences. This and the fact that the text genre is different might explain why the models trained on the Wikipedia data set does not generalise to the Tatoeba data set without a drop on performance.

The CNN uses character bi-grams as features while, with the standard settings, FastText uses only individual words to train. The better performance of the CNN might indicate that character level n-grams are more useful features for language identification than words alone.

To test this we changed the setting of FastText to train using only character level n-grams in the range 1-5 instead of individual words. In Figure~\ref{fasttextcharngram} we see the confusion matrix for this version of the FastText model. This version still achieved 97.8\% on the Wikipedia test set while improving the accuracy on the Tatoeba data set from 75.4\% to 85.8\% which is a substantial increase.

Thus, using character-level features seems to improve the FastText models' ability to generalize to sentences belonging to a domain different from the one they have been trained on.

\begin{figure}
  \centering
      \includegraphics[scale=0.5]{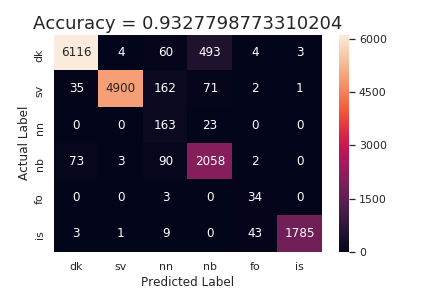}
      \caption{Results for FastText trained w. char n-grams on Wikipedia+Tatoeba and evaluated on Tatoeba.}
      \label{retrain-confuss}
\end{figure}

\begin{figure}
\centering
        \includegraphics[scale=0.5]{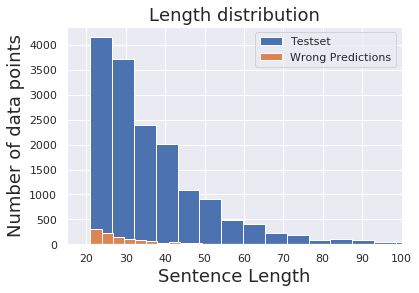}
        \caption{Distribution of sentence lengths Tatoeba test set along with the mis-classified sentences.}
        \label{retrain-lengths}
\end{figure}

\subsection{Retraining on the combined data set}
To improve the accuracy over the Tatoeba data set, we retrained the FastText model on a combined data set consisting of data points from
both the Wikipedia and Tatoeba data set.

The FastText model achieved an accuracy of 97.2\% on this combined data set and an accuracy of 93.2\% when evaluating this model on the Tatoeba test set alone - the confusion matrix can be seen in Figure~\ref{retrain-confuss}.

As was the case with the Wikipedia data set the mis-classified sentences tend to be shorter than the average sentence in the data set. In Figure~\ref{retrain-lengths} we see the distribution of sentence lengths for the Tatoeba test set along with the mis-classified sentences.

In the Tatoeba test set the mean length of sentences is 37.66 characters with a standard deviation of 17.91 while the mean length is only 29.70 characters for the mis-classified sentences with a standard deviation of 9.65. This again supports the conclusion that shorter sentences are harder to classify.

%% file: sec/07analysis.tex
\section{Analysis}

\subsection{Principal Component analysis and t-SNE}
To gain additional insight on how the different word embedding capture important information about each of the language classes, we visualized the embeddings using two different techniques for dimensionality reduction.

We used two different methods: Principal Component Analysis (PCA) and T-distributed Stochastic Neighbor Embedding (t-SNE).
We begin with a brief explanation of the two techniques and proceed with an analysis of the results.

\paragraph{Principal Component Analysis}

The first step is to calculate the covariance matrix of the dataset.
The components of the covariance matrix is given by

\begin{equation}
K_{X_i,X_j} = E[(X_i - \mu_i )(X_j -  \mu_j)]
\end{equation}

where $X_{i}$ is the $i$th component of the feature vector and $\mu_{i}$ is the mean of that component.

In matrix form we can thus write the covariance matrix as
\begin{equation}
K(\mathbf{x},\mathbf{z}) =
\begin{bmatrix}
    \text{cov}(x_1,z_1) &  \dots  & \text{cov}(x_1,z_n) \\
    \vdots & \ddots     & \vdots \\
    \text{cov}(x_n,z_1) & \dots  & \text{cov}(x_n,z_n) \\
\end{bmatrix}
\end{equation}
The next step is to calculate the eigenvectors and eigenvalues of the covariance matrix by solving the eigenvalue equation.
\begin{equation}
\det (K v-\lambda v) = 0
\end{equation}
The eigenvalues are the variances along the direction of the eigenvectors or ``Principal Components". To project our data set onto 2D space we select the two eigenvectors' largest associated eigenvalue and project our data set onto this subspace.

\begin{figure}
    \centering
    \begin{subfigure}[b]{0.47\textwidth}
      \centering

        \includegraphics[width=\textwidth]{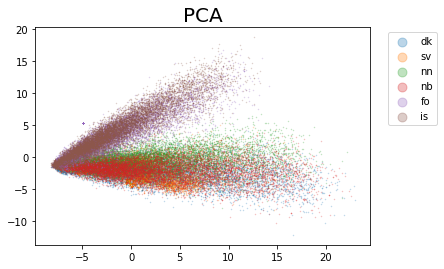}
        \caption{Character bigram}
    \end{subfigure}
    ~
    \begin{subfigure}[b]{0.47\textwidth}
          \centering
\includegraphics[width=\textwidth]{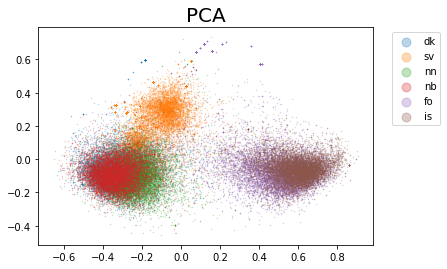}
        \caption{Fasttext cbow}
    \end{subfigure}
    ~
    \begin{subfigure}[b]{0.47\textwidth}
          \centering
\includegraphics[width=\textwidth]{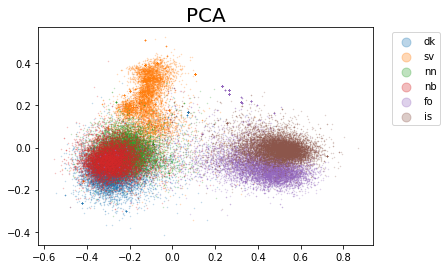}
        \caption{Fasttext skipgram}
    \end{subfigure}
    \caption{Dimensionality reduction using PCA}
    \label{pca}
\end{figure}

In Figure~\ref{pca} we see the result of running  PCA  on the wikipedia data set where we have used character level bi-grams as features, as well as the CBOW and skipgram models from FastText.

In the figure for encoding with character level bi-grams, the PCA algorithm resulted in two elongated clusters. Without giving any prior information about the language of each sentences, PCA is apparently able to discriminate between Danish, Swedish, Nynorsk and Bokmål on one side, and Faroese and Icelandic on the other, since the majority of the sentences in each language belong to either of these two clusters. \\

With the FastText implementations we observe three clusters. For both CBOW and skipgram we see a distinct cluster of Swedish sentences. When comparing the two FastText models we see that the t-SNE algorithm with skipgrams seems to be able to separate  Faroese and Icelandic data points to a high degree compared with the CBOW model. Also for the cluster identified with the sentences with Danish, Bokmål and Nynorsk the skipgram models seem seem to give a better separation, however to a lesser degree than with the two former languages.

\paragraph{t-SNE}

The T-distributed Stochastic Neighbor Embedding method was first proposed in 2008~\cite{tsne}, which favours retaining local spatial relationships over remote ones.

In t-SNE, for a given data point $x_i$, the probability of picking another data point $x_j$ as a neighbor to $x_i$ is given by:

\begin{equation}
p_{ji}= \frac{\exp (|| x_i - x_j ||^2/2\sigma_i^2 )}{\sum_{k\neq i} \exp (|| x_i - x_k ||^2/2\sigma_i^2 )}
\end{equation}

Given this probability distribution the goal is to find the low-dimensional mapping of the data points $x_i$ which we denote $y_i$ follow a similar distribution. To solve what is referred to as the ``crowding problem", t-SNE uses the Student t-distribution which is given by:

\begin{equation}
q_{ij}= \frac{ (1+|| y_i - y_j ||^2 )^{-1}}{\sum_{k\neq l} (1+|| y_k - y_l ||^2 )^{-1}}
\end{equation}

Optimization of this distribution is done using gradient decent on the Kullback-Leibler divergence which is given by:

\begin{equation}
\frac{\delta C}{\delta y_i}= 4 \sum_j (p_{ij} - q_{ij})(y_i-y_j)(1+ || y_i - y_j ||^2  )^{-1}
\end{equation}

\begin{figure}
    \centering
    \begin{subfigure}[b]{0.47\textwidth}
        \includegraphics[width=\textwidth]{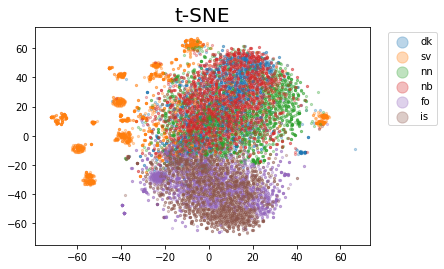}
        \caption{Character bi-gram}
    \end{subfigure}
    ~
    \begin{subfigure}[b]{0.47\textwidth}
        \includegraphics[width=\textwidth]{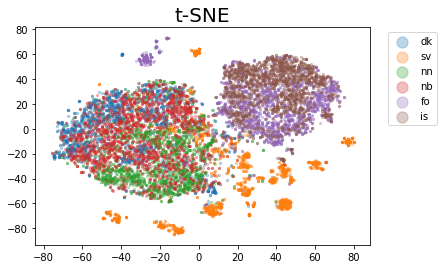}
        \caption{FastText CBOW}
    \end{subfigure}
    ~
    \begin{subfigure}[b]{0.47\textwidth}
        \includegraphics[width=\textwidth]{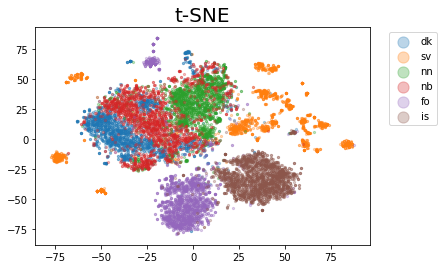}
        \caption{FastText skip-gram}
    \end{subfigure}
    \caption{Dimensionality reduction using t-SNE}
    \label{tsne}
\end{figure}

The result from running the t-SNE algorithm on the Wikipedia data set can be seen in Figure~\ref{tsne}. As was the case with PCA,  it appears that the encoding with FastText seem to capture the most relevant information to discriminate between the languages; especially the skip-gram mode does well at capturing information relevant to this task.

Here we recover some interesting information about the similarity of the languages. The data points in Bokmål lie in between those in Danish and Nynorsk, while Icelandic and Faroese have their own two clusters which are separated from the three former languages. 

This is in good agreement with what we already know about the languages. Interestingly the Swedish data points and quite scattered and the t-SNE is not able to make a coherent Swedish cluster.

This does not however mean that the Swedish data points are not close in the original space. Some care is needed when interpreting the plot since t-SNE groups together data points such that neighboring points in the input space will tend to be neighbors in the low dimensional space.

If points are separated in input space, t-SNE would like to separate them in the low dimensional space however it does not care how far they are separated. So clusters that are far away in the low dimensional space are not necessarily far away in the input space.

\subsection{Discussion}
We used the dimensionality reduction techniques PCA and t-SNE to make visualizations of feature vectors obtained by making a one-hot encoding with character bi-grams and with the two modes from FastText.

These unsupervised techniques was able to separate the sentences from Wikipedia into different clusters.
Without any prior knowledge about the actual language of each sentence these techniques indicated that the six languages can be divided into three main language categories: (1) Danish Nynorsk Bokmål (2) Faroese Icelandic and (3) Swedish.

Generally the supervised models had the largest errors when discriminating between languages belonging to either of the language groups mentioned above.

For the ``classical" models we saw that Logistic Regression and support vector machines achieved better performance than KNN and Naive Bayes, where the latter performed the worst. This was true in all cases irrespective of the method of feature extraction.

Additionally we saw that when we used feature vectors from the FastText skip-gram model the classification models achieved better results than when using either FastText CBOW or character n-grams.

Generally we saw that increasing the number of data points lead to better performance. When comparing the CNN with the ``classical" models however the CNN performed better than any of the other models even when trained on less data points. In this way it seems that the CNN achieves higher sample efficiency compared to the other models.

%% file: sec/08conclusion.tex
\section{Conclusion}
This paper presented research on the difficult task of distinguishing similar languages applied for the first time to the Scandinavian context. We describe and release a dataset and detail baseline approaches and problem analysis. 

The dataset and code are available at \url{https://github.com/renhaa/NordicDSL}.

We compared four different classical models: K nearest Neighbors, Logistic regression, Naive Bayes and a linear support vector machines with two neural network architectures: Multilayer perceptron and a convolutional neural network. The two best performing models, FastText supervised and CNN, saw low performance when going off-domain. Using character n-grams as features instead of words increased the performance for the FastText supervised classifier. By also training FastText on the Tatoeba data set as well as the Wikipedia data set resulted in an additional increase in performance.
